\documentclass[9pt,twocolumn,twoside]{pnas-new}
\templatetype{pnasresearcharticle}

\title{
Information decomposition in complex systems via machine learning
}

\author[a]{Kieran A. Murphy}
\author[a,b,c,d,e,f]{Dani S. Bassett} 

\affil[a]{Dept. of Bioengineering, School of Engineering \& Applied Science, U. of Pennsylvania, Philadelphia, PA 19104, USA}
\affil[b]{Dept. of Electrical \& Systems Engineering, School of Engineering \& Applied Science, U. of Pennsylvania, Philadelphia, PA 19104, USA}
\affil[c]{Dept. of Neurology, Perelman School of Medicine, U. of Pennsylvania, Philadelphia, PA 19104, USA}
\affil[d]{Dept. of Psychiatry, Perelman School of Medicine, U. of Pennsylvania, Philadelphia, PA 19104, USA}
\affil[e]{Dept. of Physics \& Astronomy, College of Arts \& Sciences, University of Pennsylvania, Philadelphia, PA 19104, USA}
\affil[f]{The Santa Fe Institute, Santa Fe, NM 87501, USA}

\leadauthor{Murphy} 

\significancestatement{A defining characteristic of complex systems is an abundance of variation at one scale of observation that contains, hidden within, information about organization at another scale.  To see the forest through the trees is a challenge faced whether studying society or a sandpile, climate or a brain.  We present a fully general and practical methodology, rigorously grounded in information theory, that surfaces important information out of a sea of variation in a comprehensible manner.  At its core is the concept of lossy compression: some information in a measurement is preserved and the rest is discarded.  We use machine learning to lossily compress tens to hundreds of measurements simultaneously, providing a new route to insight about complex systems through information decomposition.}

\authorcontributions{K.A.M. and D.S.B. designed research; K.A.M. performed research; K.A.M. analyzed data; and K.A.M. and D.S.B. wrote the paper.}
\authordeclaration{The authors declare no competing interests.}
\correspondingauthor{\textsuperscript{2}To whom correspondence may be addressed. E-mail: dsb@seas.upenn.edu}


\begin{abstract}
One of the fundamental steps toward understanding a complex system is identifying variation at the scale of the system's components that is most relevant to behavior on a macroscopic scale.
Mutual information provides a natural means of linking variation across scales of a system due to its independence of functional relationship between observables.
However, characterizing the manner in which information is distributed across a set of observables is computationally challenging and generally infeasible beyond a handful of measurements.  
Here we propose a practical and general methodology that uses machine learning to decompose the information contained in a set of measurements by jointly optimizing a lossy compression of each measurement.
Guided by the distributed information bottleneck as a learning objective, the information decomposition identifies the variation in the measurements of the system state most relevant to specified macroscale behavior. 
We focus our analysis on two paradigmatic complex systems: a Boolean circuit and an amorphous material undergoing plastic deformation.
In both examples, the large amount of entropy of the system state is decomposed, bit by bit, in terms of what is most related to macroscale behavior.
The identification of meaningful variation in data, with the full generality brought by information theory, is made practical for studying the connection between micro- and macroscale structure in complex systems. 
\end{abstract}

\dates{This manuscript was compiled on \today}
\doi{\url{www.pnas.org/cgi/doi/10.1073/pnas.XXXXXXXXXX}}

\begin{document}

\maketitle

\thispagestyle{firststyle}
\ifthenelse{\boolean{shortarticle}}{\ifthenelse{\boolean{singlecolumn}}{\abscontentformatted}{\abscontent}}{}

A complex system is a system of interacting components where some sense of order present at the scale of the system is not apparent, or even conceivable, from the observations of single components~\citep{nicolis2012foundations,kwapien2012complexity}.
A broad categorization, it includes many systems of relevance to our daily lives, from the economy to the internet and from the human brain to artificial neural networks~\citep{mitchell2009complexitybook,newman2011complex}.
Before attempting a reductionist description of a complex system, one must first identify variation in the system that is most relevant to emergent order at larger scales.
The notion of relevance can be formalized with information theory, wherein mutual information serves as a general measure of statistical dependence to connect variation across different scales of system behavior~\cite{matsuda2000physical,koch2018natphys}.
Information theory and complexity science have a rich history; information theory commonly forms the foundation of definitions of what it means to be complex~\cite{grassberger1986toward,tononi1994measure,gellmann1996effective,rosas2019Oinfo,golan2022pnas}.

Machine learning is well-suited for the analysis of complex systems, grounded in its natural capacity to identify patterns in high dimensional data~\cite{lecun2015deep}.
However, distilling insight from a successfully trained model is often infeasible due to a characteristic lack of interpretability of machine learning models~\citep{rudin2019stop,rudin2022interpretable}.
Restricting to simpler classes of models, for example linear combinations of observables, recovers a degree of interpretability at the expense of functional expressivity~\citep{molnar2022interpretableML}.
For the study of complex systems, such a trade-off is unacceptable if the complexity of the system is no longer faithfully represented.
In this work, we do not attempt to explain the relationship between microscale and macroscale, and are instead interested in identifying the information contained in microscale observables that is most predictive of macroscale behavior---independent of functional relationship.

We employ a recent method from interpretable machine learning that identifies the most relevant information in a set of measurements~\cite{dib_ml}.
Based on the distributed information bottleneck~\cite{aguerri2018DIB,aguerriDVIB2021}, a variant of the information bottleneck (IB)~\cite{tishbyIB2000}, the method lossily compresses a set of measurements while preserving information about a relevance quantity.
Optimization serves to decompose the information present in the measurements, providing a general-purpose method to identify the important variation in composite measurements of complex systems. 

Identifying important variation is a powerful means of analysis of complex systems, as we demonstrate on two paradigmatic examples. 
First we study a Boolean circuit, whose fully-specified joint distribution and intuitive interactions between variables facilitate understanding of the information decomposition found by the distributed IB.
Boolean circuits are networks of binary variables that interact through logic functions, serving as the building blocks of computation~\cite{savage1998models} and as elementary models of gene control networks~\cite{chaves2006methods,huynh2019gene}.
Second, we decompose the information contained in the fine-grained positional configuration of an amorphous material subjected to global deformation.
Amorphous materials are condensed matter systems composed of simple elements (e.g., atoms or grains) 
whose positional disorder gives rise to a host of complex macroscale phenomena, such as collective rearrangement events spanning a wide range of magnitudes~\citep{cubuk2017science,murphy2019transforming} and nontrivial phase transitions~\cite{jaeger1992,liu1998jammingcool,bi2011jammingshear,berthier2019gardner}.
Although the state space that describes all of the degrees of freedom is large, as is generally true of complex systems, the proposed method is able to identify the most important bits of variation by leveraging machine learning to navigate the high-dimensional space of possible lossy compression schemes.

\section{Methods} Mutual information is a measure of statistical dependence between two random variables $X$ and $Y$ that is independent of the functional transformation that relates $X$ and $Y$ (in contrast to linear correlation, for example, which measures the degree to which two variables are linearly related).
Mutual information is defined as the entropy reduction in one variable after observing the value of the other variable~\cite{cover1999elements},
\begin{equation} \label{eqn:MI}
    I(X;Y) = H(Y) - H(Y|X),
\end{equation}
with $H(X)=\mathbb{E}_{x\sim p(x)}[-\textnormal{log} \ p(x)]$ Shannon's entropy~\cite{shannon1948mathematical}.

\begin{figure}
    \centering
    \includegraphics[width=\columnwidth]{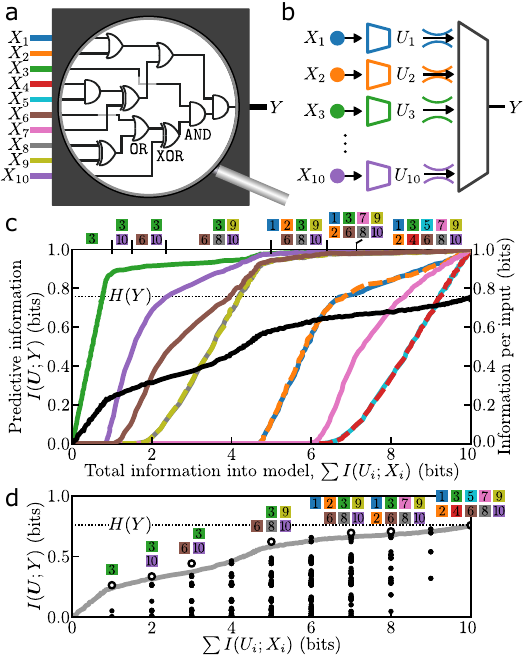}
    \caption{\textbf{Decomposing the information contained in the inputs of a Boolean circuit.} 
    \textbf{(a)}
    Ten binary inputs $\boldsymbol{X}=(X_1, ..., X_{10})$ are connected via \texttt{AND}, \texttt{OR}, and \texttt{XOR} gates to a binary output $Y$. 
    \textbf{(b)}
    A lossy compression $U_i$ is learned for each $X_i$ and then all $U_i$ are combined as input to a machine learning model trained to predict $Y$.
    \textbf{(c)}
    The distributed information plane displays the predictive information about the output (left vertical axis, black) as a function of the total information utilized about the input.
    For each value of total information into the model there is an allocation of information to the input gates indicating their relevance to the output $Y$ (right vertical axis, colors corresponding to input gates in panel \textbf{(a)}).
    The subset of inputs identified as containing the most relevant information ($I(U_i;X_i)\ge 0.1$ bits) are indicated at the top of the plot.
    Dashed lines are used for the information allocations when there is significant overlap. 
    \textbf{(d)} 
    The mutual information between all subsets of input channels 
    and $Y$ are displayed on the distributed information plane as black circles.
    The optimization of the distributed IB (gray curve) identified subsets of inputs that contain the most predictive information (open circles).
    }
    \label{fig:circuit}
\end{figure}

The information bottleneck (IB)~\cite{tishbyIB2000} is an optimization objective to extract information contained in a variable $X$ that is most relevant to another variable $Y$.
Under the IB, $X$ is lossily compressed to an auxiliary random variable $U=f(X)$ that simultaneously maximizes $I(U;Y)$ and minimizes $I(U;X)$.
Consider the case where $X$ is a composite measurement---e.g., multiple degrees of freedom of a complex system---more appropriately denoted as a random vector $\boldsymbol{X} = (X_1, ..., X_N)$.
By optimizing the IB, a lossy compression $U$ of $\boldsymbol{X}$ can be found that extracts information relevant to $Y$, but without any indication from which components the information originated.
If, instead, an information bottleneck is installed on each of the components, each $X_i$ undergoes lossy compression to a corresponding $U_i$, and then the elements of the vector $\boldsymbol{U}=(U_1, ..., U_N)$ contain the relevant information extracted from each component.
Known as the distributed IB~\cite{aguerri2018DIB,steiner2021distributedcompression}, configuring information bottlenecks in this way enables a detailed analysis of the structure of information with respect to the components~\cite{dib_ml}.
Minimization of the distributed IB Lagrangian,
\begin{equation} \label{eqn:DIB}
    \mathcal{L}_\textnormal{DIB} = \beta \sum_{i=1}^N I(U_i;X_i) - I(\boldsymbol{U};Y),
\end{equation}
extracts the entropy in $\boldsymbol{X}$ that is most descriptive of $Y$.
By sweeping over the magnitude of the bottleneck strength $\beta$, a continuous spectrum of approximations to the relationship between $\boldsymbol{X}$ and $Y$ is found, each utilizing a different amount of information from $\boldsymbol{X}$. 
The product of optimization is thus a sequence of distributed compression schemes, $\boldsymbol{U}(\beta)$, more appropriately parameterized by the total utilized information $\sum_{i=1}^N I(U_i;X_i)$.
The central focus of this work is to connect the optimized $\boldsymbol{U}$ to the structure of information contained in $\boldsymbol{X}$ for insight about the complex system under study.

In place of Eqn.~\ref{eqn:DIB}, variational bounds on mutual information have been developed that are amenable to data and machine learning~\cite{alemiVIB2016,aguerriDVIB2021}.
The lossy compression schemes are parameterized by neural networks that encode data points to probability distributions in a continuous latent space.
Stochasticity in the mapping between $X_i$ and $U_i$ allows the transmitted information to be smoothly varied and consequently facilitates optimization.  
The amount of transmitted information is upper bounded by the expectation of the Kullback-Leibler divergence~\cite{cover1999elements} between the encoded distributions and an arbitrary prior distribution, identical to the process of information restriction in a variational autoencoder~\cite{betavae,alemiVIB2016}.
While training utilizes the aforementioned upper bound, for evaluation over the course of an optimization we desire more accurate estimates of the information $I(U_i;X_i)$ extracted from each component.
We estimated upper and lower bounds derived in Ref.~\cite{poole2019variational} to obtain precise estimates of each mutual information term, where the interval between bounds was on the order of 0.01 bits.  
Although mutual information is generally difficult to estimate from data~\cite{mcallester2020infolimitations}, compressing the partial measurements $X_i$ separately segregates the information such that the amount of mutual information is small enough to allow precise estimates.
Characterization of the mutual information estimation may be found in the SI.

\section{Results}

\noindent \textbf{Boolean circuit.}
A randomly generated Boolean circuit with ten binary inputs $\boldsymbol{X}=(X_1,...,X_{10})$ and a binary output $Y$ is shown in Fig.~\ref{fig:circuit}a.
Assuming a uniform distribution over inputs, the truth table specifies the joint distribution $p(x_1,...,x_{10},y)$, and the interactions between inputs are prescribed by a wiring of logical \texttt{AND}, \texttt{OR}, and \texttt{XOR} gates. 
An information bottleneck was distributed to every input $X_i$ to monitor from where the predictive information originated via compressed variables $U_i$ (Fig.~\ref{fig:circuit}b). 
We trained a multilayer perceptron (MLP) to learn the relationship between the lossy compressions $\boldsymbol{U}$ and $Y$.

\begin{figure*}
    \centering
    \includegraphics[width=\textwidth]{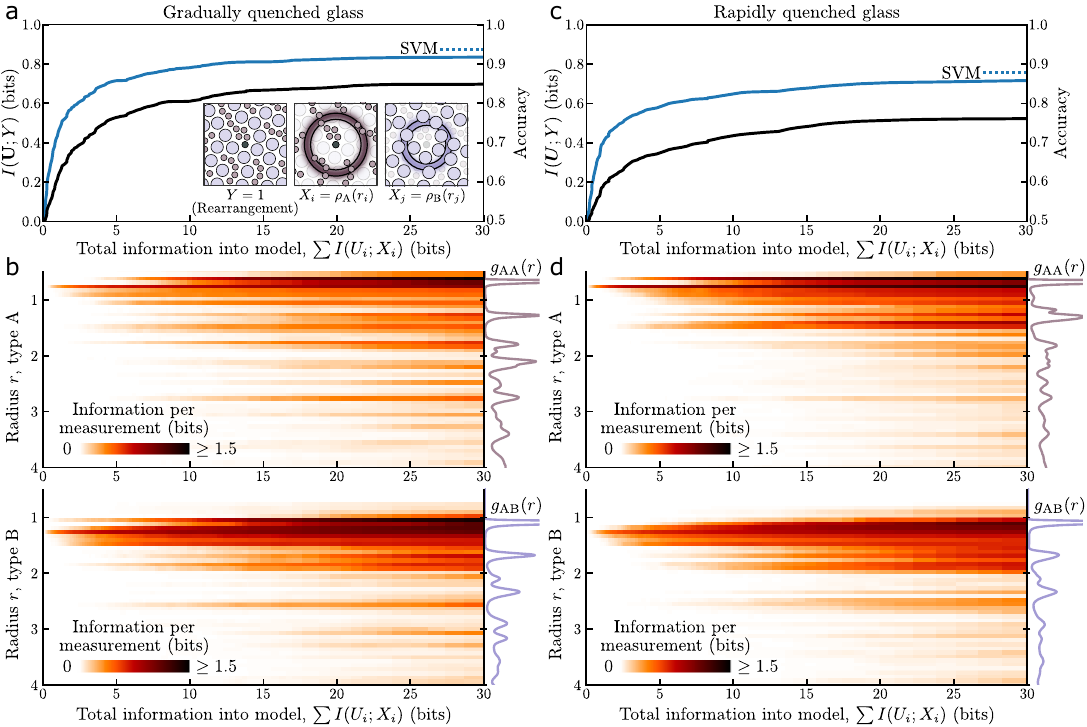}
    \caption{\textbf{Decomposing structural information about imminent rearrangement in a sheared glass.}
    \textbf{(a)} \textit{Inset:} Given a local neighborhood in a sheared glass, densities of radial shells for the small (type A) and large (type B) particles were used to predict whether the neighborhood is the locus of an imminent rearrangement event.
    \textit{Main:} For a gradually quenched glass, the information that is predictive of rearrangement (black) increased as the most predictive density information was identified and incorporated into the machine learning model.
    The accuracy (blue) was comparable to a support vector machine (SVM) (dashed line) after around twenty bits.
    \textbf{(b)} Sharing the horizontal axis with panel \textbf{(a)}, the amount of information extracted about each of the radial density measurements of small (top) and large (bottom) particles reveals the radii with the most predictive information at each level of approximation.
    The system's average density values for each particle type with type A at the center, also known as the radial distribution functions $g_\textnormal{AA}(r)$ and $g_\textnormal{AB}(r)$, are shown on the right. 
    \textbf{(c,d)} The same as panels \textbf{(a, b)} but for glass that was prepared via a rapid quench rather than a gradual quench.
}
    \label{fig:glass}
\end{figure*}

Over the course of a training run, the coefficient of the information bottleneck strength $\beta$ was swept to obtain a spectrum of compression schemes and predictive models.
The distributed information plane (Fig.~\ref{fig:circuit}c)~\citep{dib_ml} displays the predictive power as a function of the total information about the inputs $\sum I(U_i;X_i)$.
The predictive performance ranged from zero predictive information without any information about the inputs (Fig.~\ref{fig:circuit}c, lower left) to all entropy $H(Y)$ accounted for by utilizing all ten bits of input information (Fig.~\ref{fig:circuit}c, upper right). 
For every point on the spectrum there was an allocation of information over the inputs; the distributed IB objective identified the information across all inputs that was most predictive.
The most predictive information about $Y$ was found to reside in $X_3$---the input that routes through the fewest gates to $Y$---and then in the pair $X_3,X_{10}$, and so on.
Inputs that perform identical roles in the circuit were compressed nearly identically, as seen in the pairs $X_1$ and $X_2$, $X_4$ and $X_5$, and $X_8$ and $X_9$.

Machine learning facilitated the traversal of the space of lossy compression schemes of $X_i$, decomposing the information held within the circuit inputs about the output. 
Contained within the space of compression schemes are $2^{10}$ schemes that transmit complete information about discrete subsets of the inputs.
To be concrete, there are ten subsets of a single input corresponding to compression schemes that transmit one bit of information about the input gates, 45 pairs of inputs that transmit two bits, and so on, with each subset sharing mutual information with $Y$ based on the role of the specific inputs inside the circuit.
Fig.~\ref{fig:circuit}d displays the information contained in every discrete subset of inputs (black points) along with the continuous trajectory found by optimization of the distributed IB (gray curve).
The distributed IB, maximizing predictive information while minimizing information taken from the inputs, closely traced the upper boundary of discrete information allocations and identified a majority of the most informative subsets of inputs.
To decompose the information in the circuit's inputs required only a single sweep with the distributed IB, not an exhaustive search through all subsets of inputs.
We note that the product of the distributed IB is not an ordering of single variable mutual information terms $I(X_i;Y)$, which would be straightforward to calculate, but instead the ordering of information selected from all of $\boldsymbol{X}$ that is maximally informative about $Y$.
Analysis of additional Boolean circuits with the distributed IB as well as comparisons to feature importance derived from logistic regression and Shapley values~\cite{molnar2022interpretableML,shap,sage} may be found in the SI.

\noindent \textbf{Decomposing structural information in a physical system.}
Linking structure and dynamics in amorphous materials---complex systems consisting of particles crowded together in a disordered configuration---has been a longstanding challenge in physics~\cite{argon1979bubbles,falk1998dynamics,manning2011vibrational}. 
In particular, the deformation of amorphous materials generally proceeds in fits and starts, punctuated by sudden, localized rearrangement events reminiscent of earthquakes along a tectonic fault~\cite{dahmen2011simple,murphy2019transforming,richard2020indicators}.
Where, in the disordered configuration of the particles, is the information about whether a sudden rearrangement is about to occur?
Searching for signatures of collective behavior in the multitude of microscopic degrees of freedom is an endeavor emblematic of the study of complex systems more generally and one well-suited for machine learning and information theory.
We accept that the functional relationship between the micro- and macroscale variation is potentially incomprehensible, and are instead interested in the information at the microscale that is maximally predictive of behavior at the macroscale.
While prior work has analyzed the information content of hand-crafted structural descriptors individually~\cite{dunleavy2012information,jack2014information,dunleavy2015mutual}, the distributed IB searches through the space of information from many structural measurements in combination. 

Two-dimensional simulated glasses, prepared by either rapid or gradual quenching and composed of small (type A) and large (type B) particles that interact with a Lennard-Jones potential, were subjected to global shear deformation~\cite{richard2020indicators}.
For each event, the location that precipitated the rearrangement was identified and paired with negative samples from elsewhere in the system to create a binary classification dataset.

We first characterized the microscale structure with a scheme of measurements that has been associated with plastic rearrangement in a variety of amorphous systems: the densities of radial bands around the center of a region~\cite{behler2007structurefns,cubuk2015PRL}.
By training a support vector machine (SVM) to predict rearrangement based on the radial density measurements, a linear combination of the values is learned. In the literature, that combination is commonly referred to as \textit{softness}, and has proven to be a useful local order parameter~\cite{schoenholz2016natphys,softnessGrainBoundaries,softnessFilms,ridout2021avalanche}.

We approached the same prediction task from an information theoretic perspective, seeking the specific bits of variation in the density measurements that are most predictive of collective rearrangement.
Each radial density measurement underwent lossy compression by its own neural network before all compressions were concatenated and used as input to an MLP to predict rearrangement.
By sweeping $\beta$, a single optimization recovered a sequence of distributed compression schemes, each allocating a limited amount of information across the 100 density measurements to be most predictive of imminent rearrangement (Fig.~\ref{fig:glass}).

The trajectories in the distributed information plane (Fig.~\ref{fig:glass}a,c), for both gradually and rapidly quenched glasses, reflect the growth of predictive information and prediction accuracy given maximally predictive information about the radial densities.
With only one bit of information from the density measurements, 71.8\% predictive accuracy was achieved for the gradually quenched glass and 69.5\%  was achieved for the rapidly quenched glass; with twenty bits, the accuracy jumped to 91.3\% and 85.4\%, respectively.
Beyond twenty bits of density information, the predictive accuracy became comparable to that of the support vector machine, which can utilize all of the continuous-valued density measurements for prediction with a linear relationship.

For every point along the trajectory, information was identified from the density measurements that, together, formed the combination of bits that were most predictive of rearrangement. 
The allocation of information across the one hundred radial density measurements would be unintelligible if displayed in the distributed information plane as in Fig.~\ref{fig:circuit}c; instead we employ heatmaps where each row corresponds to a different density measurement (Fig.~\ref{fig:glass}b,d).
The majority of the information was selected from smaller radii (close to the center of the region), which can be expected given the localized nature of rearrangement events \cite{argon1979bubbles,falk1998dynamics}.
Less intuitive is the information decomposition as it relates to the radial distribution functions $g_\textnormal{AA}(r)$ and $g_\textnormal{AB}(r)$, the system-averaged radial densities of type A and B particles in regions with a type A particle at the center.
For both glasses, the most predictive bits originated in the low density radial bands nearest the center.
As more information was incorporated into the prediction (moving left to right in the heatmaps), the additional bits came from radial bands that corresponded to particular features of $g_\textnormal{AA}(r)$ and $g_\textnormal{AB}(r)$.
Outside of the first low density region, the selected information often came from the high density radii of type A particles and the low density radii of type B particles; this trend held true for both glasses.
While the information decomposition highlighted similar features in both glasses, the more pronounced structure of selected information out to larger radii for the gradually quenched glass is indicative of its higher structural regularity, which is also seen in the pronounced features of its radial distribution functions $g_\textnormal{AA}(r)$ and $g_\textnormal{AB}(r)$.
For comparison, the weights of the trained SVM and estimated Shapley values may be found in the SI.

\begin{figure}
    \centering
    \includegraphics[width=\columnwidth]{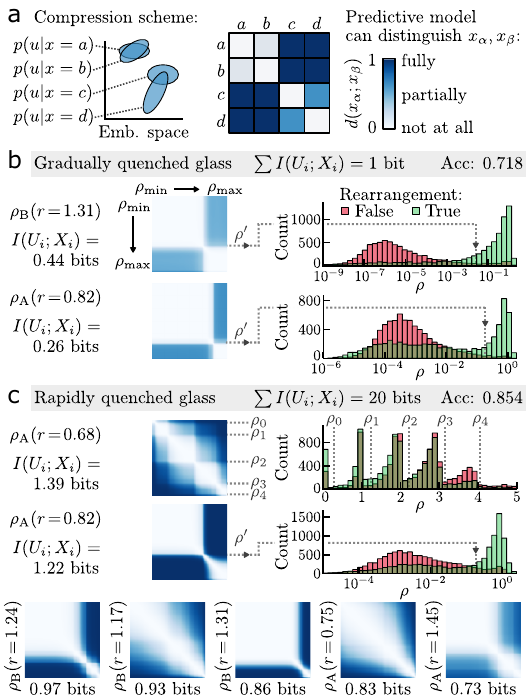}
    \caption{\textbf{Selected bits of information as distinctions among raw measurement values.}
    \textbf{(a)} 
    Lossy compression is achieved by mapping the raw values of $X$ to probability distributions in latent space.
    The statistical similarity of the conditional distributions, visualized as a distance matrix for all pairs of feature values, determines how distinguishable the raw feature values are to the predictive model.
    \textbf{(b)}
    The single most predictive bit of information about rearrangement in the gradually quenched glass came predominantly from two density measurements.
    The distinguishability matrices indicate that the compression scheme applied a simple threshold to these measurements: density values less than a cutoff value $\rho^\prime$ were indistinguishable from each other, as were values above the cutoff.
    The histograms of density values conditioned on rearrangement (right) show that the learned cutoff value separates the probability masses.
    \textbf{(c)}
    The twenty most predictive bits of radial density information in the rapidly quenched glass were selected from many radial bands.
    The two that contribute more than a bit of information each correspond to the density of type A particles near the center; one compression scheme effectively counted the number of particles in the high density shell.
    The distinguishability matrices of the next five most informative radial bands are shown below.
}
    \label{fig:glass_compression_schemes}
\end{figure}

The amount of information extracted from each density measurement was predominantly a single bit or less.
In contrast to the Boolean variables in Fig.~\ref{fig:circuit}c, where $I(U_i;X_i)$ completely characterized the compression scheme, a continuous-valued density can be compressed to a bit in any number of ways. What was the specific variation extracted from each density measurement?
Through inspection of the learned compression schemes, the extracted information can be further decomposed by the degree of distinctions between measurement values that were conveyed to the predictive model (Fig.~\ref{fig:glass_compression_schemes}a)~\cite{dib_ml}.
Serving to visualize a compression scheme, a distinguishability matrix is a distance matrix between values $x_\alpha$ and $x_\beta$ where a value of 0 indicates the states are indistinguishable, a value of 1 means that they are fully distinguishable, and an intermediate value means they are partially distinguishable.  
We found that the single most important bit of information for the gradually quenched glass (Fig.~\ref{fig:glass_compression_schemes}b) was a composition of partial bits from multiple density measurements, mostly from the first low-density shell of each type of particle.
For both measurements, the compression scheme acted as a threshold on the range of possible density values: values less than a cutoff $\rho^\prime$ were indistinguishable from each other for the purposes of prediction and were partially distinguishable from density values above the cutoff.
By examining the distribution of density values in these radial shells, we see that the cutoff values leverage the separability of the density distributions when conditioned on rearrangement. 

With more information utilized for prediction, some of the compression schemes differed from simple thresholds (shown for the rapidly quenched glass in Fig.~\ref{fig:glass_compression_schemes}c).
For the predictive model operating with a total of twenty bits of density information, two density measurements contributed more than a bit each.
The learned compression of the first high-density shell of type A particles essentially counted the number of particles in the shell, with distinguishability between densities as if there were several thresholds over the range of the values that act to roughly discretize the density measurement.

\begin{figure}
    \centering
    \includegraphics[width=\columnwidth]{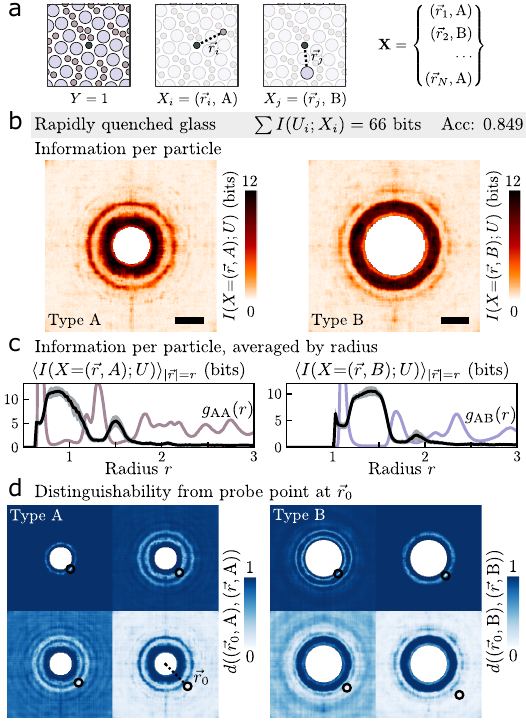}
    \caption{\textbf{Information decomposed in terms of per-particle measurement basis.}
    \textbf{(a)}
    Instead of the density of radial shells, each particle's position and type in a local neighborhood were used as input measurements to relate to rearrangement.
    \textbf{(b)}
    The per-particle information transmitted as a function of particle position, for the small type A (left) and large type B (right) particles, for the predictive model utilizing 66 bits of information about the rapidly quenched glass. 
    The scale bar is a distance of one in simulation units, equal to the length scale of interaction between types A and B particles.
    \textbf{(c)}
    Averaged radially, the information (black) resides in particles that are situated in the first troughs of the radial distribution function, $g(r)$ (colored curves).
    \textbf{(d)} 
    For a particle at position $\vec{r}_0$, the distinguishability of particles of the same type at all other locations has a radial structure and indicates that negligible azimuthal information was transmitted. 
}
    \label{fig:settransformer}
\end{figure}

Decomposing the information in a composite $\boldsymbol{X}$ depends upon its basis of measurements~\cite{dib_orig}.
In the study of complex systems, there can be multiple `natural' schemes of measuring a system state.
Measurements of the densities of radial bands lead to an essentially linear relationship between structure and rearrangement~\cite{schoenholz2016natphys}; what if we had not inherited such a fortuitous measurement scheme?
Another natural basis of measurements is the position of all of the particles (Fig.~\ref{fig:settransformer}a).
To be clear, instead of representing a system's local configuration in terms of one hundred measurements of radial density, we could have directly used the $N$ measurements of the position and type of the $N$ particles in a local region.
In the preceding analysis, each radial density was considered a distinct information source and a unique compression channel was learned for each.
In the per-particle measurement basis, there is no straightforward means to divide the measurements into separate sources because positions exist along a continuum; accordingly, a single compression channel was used for each type, compressing in parallel every particle of type A and then similarly for type B.
To subsequently predict rearrangement based on the set of compressed per-particle measurements, which lack a canonical ordering, we used a permutation-invariant transformer architecture.~\citep{lee2019set}.

The per-particle measurement scheme imposed no positional structure on the selection of configurational information. 
Nevertheless, we found that the information cost per particle as a function of the position in the neighborhood had a radial structure (Fig.~\ref{fig:settransformer}b).
The information per particle was highest in the low density radial bands near the center of the region (Fig.~\ref{fig:settransformer}c), and inspection of the compression scheme indicated that negligible azimuthal information was transmitted (Fig.~\ref{fig:settransformer}d).
The information decomposition allowed for similar insights to be derived as in the radial density measurement scheme, even though the nature of the predictive model in the two cases was substantially different.
Additionally, because the distributed IB operates entirely on the input side of an arbitrary predictive model, the information analysis was agnostic to whether the model was a simple fully connected network or a more complicated transformer architecture.

\section{Discussion}
A universal challenge faced when studying complex systems, fundamental to what makes a system \textit{complex}, is the abundance of entropy from the perspective of the microscale that obscures relevant information about macroscale behavior. 
The generality of mutual information as a measure of statistical relatedness, and the capacity of deep learning to handle high-dimensional data, allow the distributed IB to be as readily utilized to identify structural defects relevant to a given material property as it is to reveal gene variation relevant to a given affliction.  
Tens, hundreds, and potentially thousands of measurements of a complex system are handled simultaneously, rendering practical analyses that would have previously been infeasible through exhaustive search or severely limited by constraints on functional relationships between variables.  

Information theory has long held appeal for the analysis of complex systems owing to the generality of mutual information~\cite{cover1999elements,nicolis2012foundations}.
However, the estimation of mutual information from data is fraught with difficulties~\cite{mcallester2020infolimitations,saxe2019,poole2019variational}, and the rapid growth in ways information can be distributed across a number of variables~\cite{williams2010PID} have hindered information theoretic analyses of data from complex systems.
By distributing information bottlenecks across multiple partial measurements of a complex system, entropy is partitioned to a degree that makes precise estimation of mutual information possible while simultaneously revealing the most important combinations of bits for insight about the system.
Machine learning navigates the space of lossy compression schemes for each variable and allows the identification of meaningful variation without consideration of the black box functional relationship found by the predictive model.

Instead of compressing partial measurements in parallel, the information bottleneck~\cite{tishbyIB2000} extracts the relevant information from one random variable in its entirety about another, and is foundational to many works in representation learning~\cite{zaidi2020IBreview,goldfeld2020information}.
In the physical sciences, the IB has been used to extract relevant degrees of freedom with a theoretical equivalence to coarse-graining in the renormalization group~\cite{gordonrelevance2021,kline2021RGIB}, and to identify useful reaction coordinates in biomolecular reactions~\cite{wang2019PIB}.
However, the IB generally has limited interpretability because the singular bottleneck occurs after processing the complete input, allowing the compression scheme to involve arbitrarily complex relationships between components of the input without penalty. 
Additionally, when the relationship between $\boldsymbol{X}$ and $Y$ is deterministic (or nearly so), the entire spectrum of extracted information is trivially a copy of $Y$ with added noise~\cite{kolchinsky2018caveats,kolchinskyNonlinearIB2019,dib_orig}.
The distribution of information bottlenecks across observables is critical to an interpretable information decomposition and to recovering an illuminating spectrum of extracted information.

A growing body of literature focuses on a fundamentally different route to decompose the information contained in multiple random variables $\{X_i\}$ about a relevant random variable $Y$; that alternative route is partial information decomposition (PID) \cite{williams2010PID,gutknecht2021bits}. 
Although there is no consensus on how to achieve PID in practice, its goal is to account for the mutual information between $\{X_i\}$ and $Y$ in terms of subsets of $\{X_i\}$, by analogy to set theory~\citep{kolchinsky2022PID}.
PID allocates information to the input variables in their entirety, whereas the distributed IB selects partial entropy from the input variables in the form of lossy compression schemes, with one scheme per variable.
While PID has been proposed as an information theoretic route to study complex systems~\cite{varley2022emergence} and quantify complexity~\cite{ehrlich2022partial}, the super-exponential growth of PID terms renders the methodology somewhat impractical unless ``coarse-grained'' metrics are used~\cite{timme2014SynRedReview,rosas2019Oinfo}.
There are $5\times 10^{22}$ PID terms for a Boolean circuit with 8 inputs~\cite{williams2010PID} and the number of terms for the 10-input circuit from Fig.~\ref{fig:circuit} is not known~\cite{gutknecht2021bits}.
By contrast, the distributed IB offers a pragmatic route to the decomposition of information in a complex system: it is amenable to machine learning and data, and can readily process one hundred (continuous) input variables as in the amorphous plasticity experiments.
Although the distributed IB navigates a different space of information terms than those induced by PID, it may prove fruitful to employ the former as a means to focus on the most significant terms of the latter.

\subsection*{Data, Materials, and Software Availability} The glass dataset is attached to this paper and has also been deposited to Figshare~\cite{glass_dataset}. The code base, including code to generate truth tables from Boolean circuits, has been posted to Github and may be found through the following link: \href{https://distributed-information-bottleneck.github.io}{distributed-information-bottleneck.github.io}.

\subsection*{Acknowledgements}
We gratefully acknowledge Sam Dillavou and Zhuowen Yin for helpful discussions and comments on the manuscript, and Sylvain Patinet for the amorphous plasticity data.


\clearpage

\onecolumn

\setcounter{section}{0}
\setcounter{page}{1}
\setcounter{figure}{0}
\setcounter{table}{0}

\renewcommand{\thepage}{S\arabic{page}}
\renewcommand{\thesection}{S\arabic{section}}
\renewcommand{\thetable}{S\arabic{table}}
\renewcommand{\thefigure}{S\arabic{figure}}
\renewcommand{\figurename}{Supplemental Material, Figure}

\hrule
\vspace{3mm}

{\Large Supplemental Material}

\subsection*{Code availability}
The full code base has been released on Github and may be found through the following link: \href{https://distributed-information-bottleneck.github.io}{distributed-information-bottleneck.github.io}.
Every analysis included in this work can be repeated from scratch with the corresponding Google Colab iPython notebook in \href{https://github.com/distributed-information-bottleneck/distributed-information-bottleneck.github.io/tree/main/complex_systems}{this directory}.

\section*{Boolean circuitry: Comparative analyses and extended examples}
To further develop intuition about the manner of information decomposition achieved by the distributed IB, we analyze several additional Boolean circuits in this section.  
As a reminder, the distributed IB ingests a dataset of input-output observations and yields a decomposition of the information contained in the inputs about the output.
The decomposition offers a degree of interpretability about the relationships between the inputs so far as they determine the output.
The ground truth circuitry in these examples is used to create the data but then does not factor into the information decomposition; it is displayed only as a point of reference.

\begin{figure*}
    \centering
    \includegraphics[width=\textwidth]{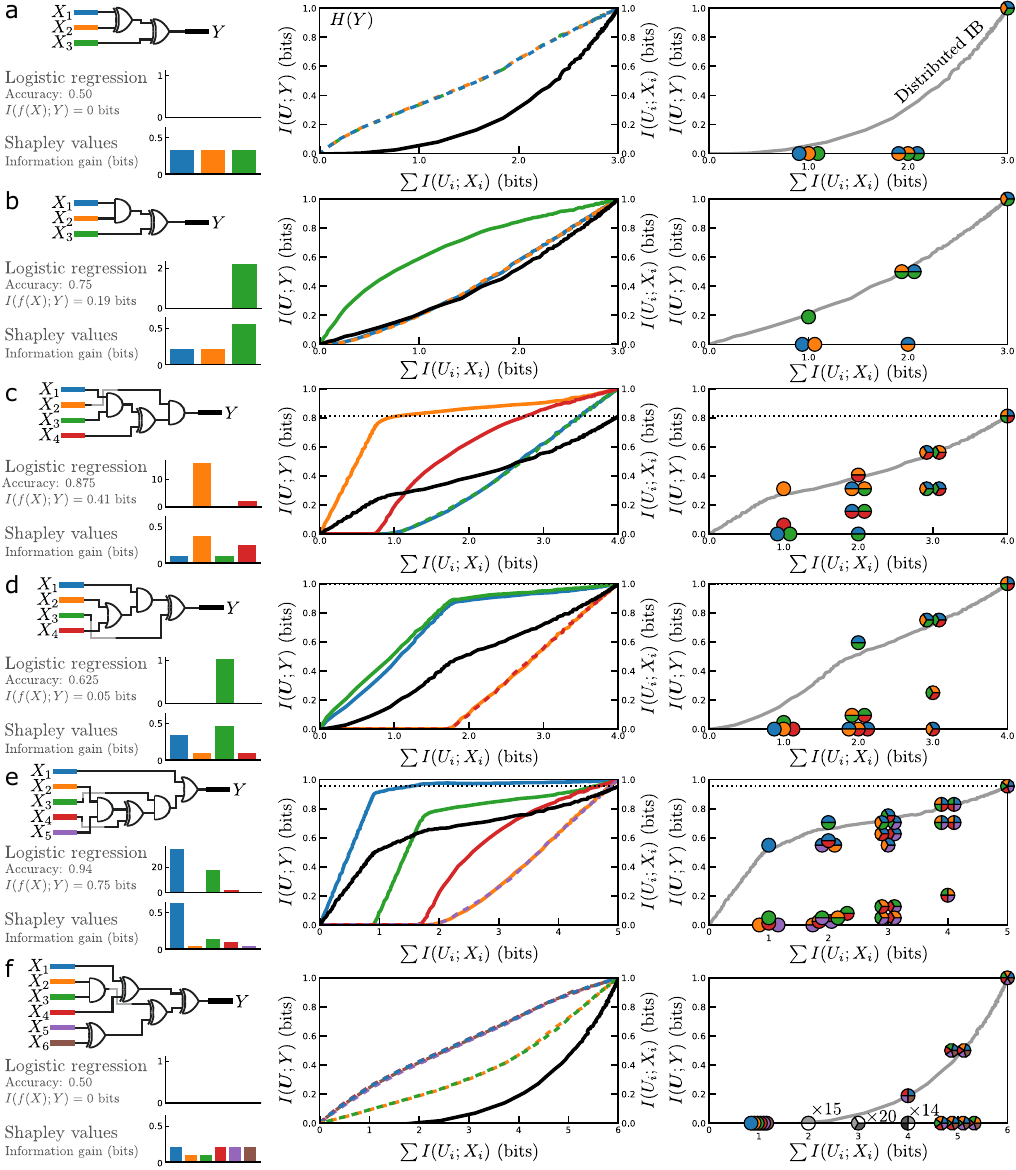}
    \caption{\textbf{Information decomposition of additional Boolean circuits.}
    \textbf{(a-f)} Distributed IB analysis of randomly generated Boolean circuits, from three to six input gates.
    The circuit diagram (left) displays the circuit that was used to generate the joint distribution for training the distributed IB.
    Under each circuit we show two alternative routes to probe the importance of $X_i$ with respect to $Y$: model weights of logistic regression and Shapley values regarding information gain about the output $Y$.
    The distributed information plane (middle) shows the total information $I(\boldsymbol{U};Y)$ (black) and the information allocation by input gate $I(U_i;X_i)$ (color) as a function of total information transmitted to the predictive model, $\sum I(U_i;X_i)$.
    Information allocation curves are dashed in cases where it aids visual clarity.
    On the right, we reproduce the distributed IB trajectory in gray and compare to the information contained by each of the possible discrete subsets.
    Each discrete subset is colored according to the input gates in the subset.
    We have shifted points horizontally when there was overlap.
    In all plots, the horizontal dashed line indicates $H(Y)$, the entropy of $Y$, though it occasionally coincides with the upper horizontal axis.
    In \textbf{f}, we suppress visualization of the subsets that all have zero information, and instead indicate the number of such subsets.
    }
    \label{fig:bool_extended}
\end{figure*}

First, we randomly generated Boolean circuits with three to six input gates (Fig.~\ref{fig:bool_extended}).
To serve as reference analyses, we analyzed data from the circuits with a linear and a nonlinear method that each provide a sense of the importance of the inputs $X_i$ in determining the output $Y$.  
We performed logistic regression and computed the Shapley values in relation to the mutual information between $\{X_i\}$ and $Y$~\cite{shap,sage,molnar2022interpretableML} (Fig.~\ref{fig:bool_extended}, left column).
We then optimized the distributed IB following the training protocol of Fig.~\ref{fig:circuit} (Fig.~\ref{fig:bool_extended}, middle column).
We again compared the distributed IB decomposition to the exhaustive set of mutual information terms, where there is one for each possible subset of inputs (Fig.~\ref{fig:bool_extended}, right column). 
By contrast to Fig.~\ref{fig:circuit} of the main text, the circuits are small enough to allow visualization of the specific input combination (shown as pie charts) represented by each point in the plot.

In \textbf{logistic regression}, a linear combination of the inputs is fit in log-probability space to be maximally predictive of the label $Y$, and this linearity grants interpretability to the model weights.
However, the linearity also severely restricts the relationships that can be modeled.
In the six circuits in Fig.~\ref{fig:bool_extended}, the nonlinearity of \texttt{XOR} gates diminished predictive power of the logistic models: the amount of information captured by the model, $I(f(X);Y)$, was less than $0.2$ bits for four of the circuits. 
In the best case (the circuit in panel (e), with only one out of four gates being \texttt{XOR}), the model captured 0.75 bits of information about $Y$ and had an accuracy of 94\%.
Despite the inability to represent the nonlinearity of \texttt{XOR}, there was some degree of success in shedding light on the inner workings of the circuit: the model coefficients revealed one or a few of the most informative inputs in circuits (b), (c), (d), and (e).

\textbf{Shapley values}~\cite{molnar2022interpretableML} have become a popular approach to explaining machine learning models because they convey a sense of importance of inputs (or features) with regard to an output while being agnostic to the model used.  
Originating in game theory as a way to assign credit to a set of players in a game, the most common formulation in explainable ML~\cite{shap} grants \textit{local} interpretability~\cite{molnar2022interpretableML} to a trained model; i.e., it provides case-by-case explanations of model outputs for individual examples from the data.
By contrast, the distributed IB provides \textit{global} interpretability~\cite{dib_ml}, meaning the insights pertain to the entire data distribution (and the entire system) at once.  
Instead of assigning credit for a single model prediction, Shapley values can be computed with regard to the information gain of a model in order to provide global interpretability~\cite{sage}.
We use the truth table for each circuit as a perfect model, though we could also have trained a neural network and estimated Shapley values following Covert \emph{et al.}~\cite{sage}.
The Shapley value for each input is a weighted sum of all possible mutual information terms (the same terms displayed in the right column of Fig.~\ref{fig:bool_extended}), serving to summarize the contribution of information for each input in the context of every combination of other inputs.
A desirable property of Shapley values is that they are additive: for the deterministic relationships represented by these circuits, the Shapley values for all inputs sum to the entropy $H(Y)$ of the output.

The Shapley values relative to information gain conveyed more about the underlying circuitry than the logistic model coefficients, which is sensible given that the ``model'' used for the Shapley values is the full truth table.
We found that the values conveyed the same high-level feature importance as the distributed IB, successfully accounting for higher-order interaction effects between the gates because the information in all subsets of inputs is included in the calculation.

In contrast to logistic regression and Shapley values, which produce a single value of importance per input, the distributed IB produces an entire spectrum of importance values in the form of information allocations across the inputs.
The spectrum represents a Pareto frontier, where every point in the spectrum contains the most information about the output $Y$ for the least total information about the inputs $X_i$.
The allocations order the information contained in the inputs from most to least relevant about the output $Y$.
By reading the distributed information plane from left to right (middle column of Fig.~\ref{fig:bool_extended}), we infer the importance of the inputs by the order in which they appear in the information allocations.
The rank order of the importance of the inputs matches that of the Shapley values in every circuit.
What else can be learned from the distributed IB by way of the spectrum of information allocations?

Consider the three-input circuits of Fig.~\ref{fig:bool_extended} (a) and (b).
Given only the Shapley values for the three inputs in panel (a) we infer that the inputs are equally responsible for determining the output, but little else is learned. 
With the information decomposition from the distributed IB, we learn far more about the circuit.
The identical trajectories of the inputs in the distributed information plane suggests the inputs perform equivalent roles.
The slow growth of the predictive information $I(\boldsymbol{U};Y)$ suggests a highly entangled interaction between the inputs whereby substantial information about all three is required before any information is gleaned about $Y$.
By comparison, the growth of predictive information for the circuit in panel (b), which increases immediately with information about $X_3$, tells us that information about $Y$ is available with information about only the third input.

Consider again the three curves displaying the information allocation to the inputs in the distributed information plane of Fig.~\ref{fig:bool_extended}b.
One (corresponding to $X_3$) is concave while the other two ($X_1$ and $X_2$) are convex and mirror each other; the curves tell about the growth of information of the inputs in combination.
Similar behavior can be seen in Fig.~\ref{fig:bool_extended}c with $X_4$, $X_1$ and $X_3$; inspection of the circuits shows there is an \texttt{AND}-\texttt{XOR} subcircuit common to the circuits in both panels (b) and (c).
By chance (because the circuits were randomly sampled) the replication can be taken one step further.
The entire information allocation of the circuit in Fig.~\ref{fig:bool_extended}c is found in the circuit in Fig.~\ref{fig:bool_extended}e, and again the circuits can be observed to share a subcircuit.
The information allocations and the growth of predictive information tell much more about the interactions between inputs than can be conveyed in a single list of importance values.

In the right column of Fig.~\ref{fig:bool_extended}, we display the information about $Y$ contained in all possible subsets of inputs, allowing a check on the quality of information allocations recovered by the distributed IB.
As in the main text, we again found that the distributed IB was able to identify the most informative subsets without requiring exhaustive search.
Notably, for a circuit of only \texttt{XOR} gates (Fig.~\ref{fig:bool_extended}a), the distributed IB outperformed the discrete allocations by leveraging partial information allocations.
With only discrete information allocations across the inputs, there is a discontinuous transition from zero information about $Y$ with knowledge of any two input gates, to complete information about $Y$ when knowing all three.
In stark contrast, by exploring the larger space of soft compression schemes (in which partial information can be transmitted), machine learning found a smooth interpolation between zero and complete information about $Y$.

\begin{figure*}
    \centering
    \includegraphics[width=\textwidth]{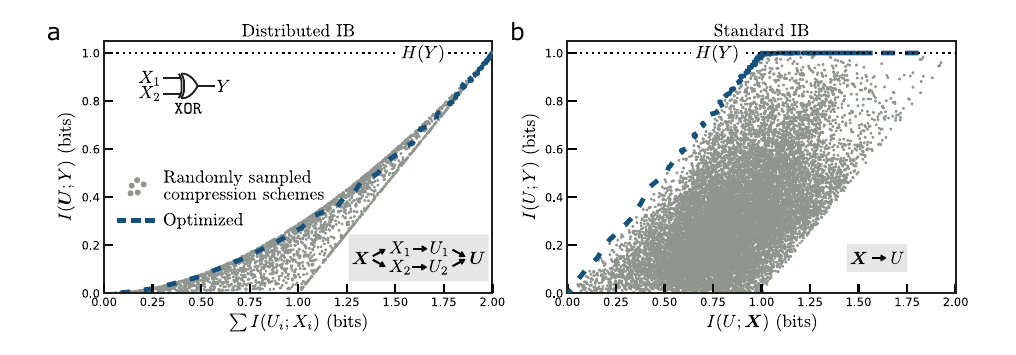}
    \caption{\textbf{The space of compression schemes navigated by the distributed and standard IB.}
    \textbf{(a)} For a single \texttt{XOR} gate, the two inputs are compressed independently according to the distributed IB, in terms of relevance to the output $Y$. 
    We randomly sample 2500 compression schemes (gray points) and additionally optimize the variational formulation of the distributed IB from the main text (blue, dashed).
    \textbf{(b)} With the same \texttt{XOR} gate as in \textbf{a}, the full input $\boldsymbol{X}$ is compressed to a single compression variable $U$. 
    We randomly sample 15000 compression schemes (gray points) and additionally optimize the variational IB (blue, dashed).
    The space of distributed compression schemes in \textbf{a} is contained within the space of compression schemes of \textbf{b}. 
    Because the components of $\boldsymbol{X}$ are independent in this example, the horizontal axes of \textbf{a} and \textbf{b} are equivalent.
    }
    \label{fig:xor_analysis}
\end{figure*}

\subsection*{Comparison of the space of compression schemes relevant to the distributed and standard IB}
In Fig.~\ref{fig:xor_analysis} we analyzed a single \texttt{XOR} logic gate with both the distributed IB and the standard IB.
For the distributed IB (Fig.~\ref{fig:xor_analysis}a), each of the inputs $X_1$ and $X_2$ was compressed to its own variable, $U_1$ and $U_2$, defined by the conditional distributions $p(u_1|x_1)$ and $p(u_2|x_2)$.
We sampled from the space of distributed compression schemes by randomly sampling conditional distributions over a two-symbol alphabet for each $U_i$.
2500 distributed compression schemes are plotted as gray dots in the distributed information plane in Fig.~\ref{fig:xor_analysis}a, along with the trajectory navigated by the distributed IB.
The convex shape of the distributed IB trajectory closely traced the boundary of distributed compression schemes.

By contrast to the distributed compression schemes, the standard IB compresses the entire input $\boldsymbol{X}$ to a single variable $U$ and optimizes the IB Lagrangian~\cite{tishbyIB2000}
\begin{equation}
    \mathcal{L}_\textnormal{IB} = \beta I(U;\boldsymbol{X}) - I(U;Y).
\end{equation}
We sampled from the space of compression schemes by randomly sampling conditional distributions over a four-symbol alphabet for $U$ (Fig.~\ref{fig:xor_analysis}b).
In total, 15000 compression schemes were created by sampling the sixteen values of the relevant conditional distributions from a uniform distribution and raising to the sixth power before normalizing.
We found that sampling probability values uniformly (i.e., without exponentiation) rarely sampled compression schemes with $I(U;\boldsymbol{X})\gtrsim 1$ bit.
With the same variational bounds we used for the distributed IB~\cite{alemiVIB2016}, we optimized the standard IB and found that it successfully traced the boundary of relevant compression schemes.

\begin{figure*}
    \centering
    \includegraphics[width=\textwidth]{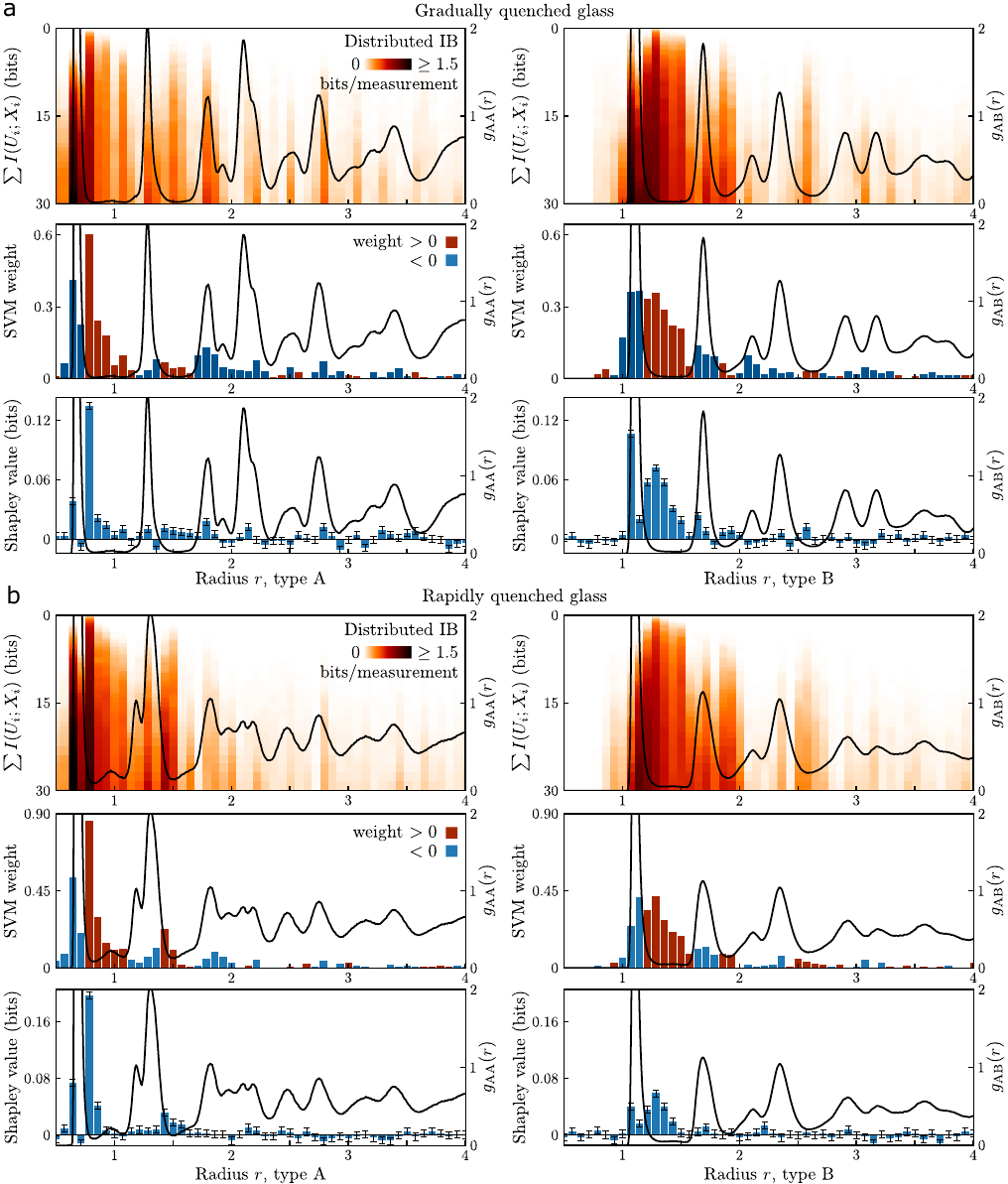}
    \caption{\textbf{Comparative analyses of radial density measurement scheme.}
    \textbf{(a)} For the simulated glass prepared with a gradual quench, we reproduce the information allocation from the main text and compare to the weights of a support vector machine (SVM) and to the estimated Shapley values for the model's information gain. 
    The absolute value of the SVM weights are shown, with red (blue) to indicate that the weight is positive (negative).
    The error bars on the Shapley values are estimated during sampling according to SAGE~\cite{sage}. 
    The radial densities for type A particles are shown on the left and type B on the right, though we note that both sets of densities were used in combination as input to the models.
    The radial distribution functions $g_\textnormal{AA}(r)$ and $g_\textnormal{AB}(r)$, showing system-averaged radial densities, are the black curves and utilize the right vertical axes. 
    \textbf{(b)} The same as \textbf{a}, for the rapidly quenched glass.
    }
    \label{fig:glass_comparison}
\end{figure*}

The space of distributed compression schemes is more restrictive than the space of compression schemes relevant to the standard IB. 
Because the components of $\boldsymbol{X}$ are independent for this example, the two horizontal axes of Fig.~\ref{fig:xor_analysis}a,b are equivalent, allowing the point scatter of compression schemes to be directly compared.  
While the distributed schemes in Fig.~\ref{fig:xor_analysis}a are included in the space of schemes in Fig.~\ref{fig:xor_analysis}b, only the latter includes schemes that utilize knowledge of the full state of $\boldsymbol{X}$.
As an example, one scheme that optimizes the standard IB maps $\boldsymbol{X} \in \{01, 10\}$ and $\boldsymbol{X} \in \{00, 11\}$ to two separate clusters, effectively extracting $Y$ without revealing anything about the structure of $\boldsymbol{X}$---a shortcoming of the standard IB when analyzing deterministic relationships~\cite{kolchinsky2018caveats}.  

\section*{Glassy rearrangement: Comparative analyses of radial density measurements}

For the simple Boolean circuits of Fig.~\ref{fig:bool_extended}, all three analyses---logistic regression, Shapley values, and the distributed IB---conveyed a sense of importance of $X_i$ in determining $Y$.
The unique capabilities of the distributed IB become more apparent under more challenging scenarios, such as relating rearrangement in a simulated glass to local radial density measurements.
In Fig.~\ref{fig:glass_comparison}, we reproduce from Fig.~\ref{fig:glass} of the main text the distributed IB information decomposition and compare it to the weights of a support vector machine (SVM) as well as the Shapley values estimated for an MLP\footnote{Two layers of 256 \texttt{Leaky ReLU} units, trained for 10 epochs.} (following a method named SAGE~\cite{sage}\footnote{\href{https://github.com/iancovert/sage}{https://github.com/iancovert/sage}}).
There is general agreement about the most important radial bands as told by the order of information allocation by the distributed IB, and the magnitude of weights derived from the (linear) SVM and the (nonlinear) Shapley values.
The distributed IB reveals far more, however: all of the most informative subsets of radial bands and the specific bits of information from each radial density that are relevant to rearrangement (Fig.~\ref{fig:glass_compression_schemes} of the main text).
To find the most informative subsets of radial bands, the authors of~\cite{schoenholz2016natphys} trained millions of SVMs, each on different subsets of the radial densities---quickly making exhaustive evaluation impractical beyond a handful of measurements.

Regarding the specific bits of relevant information (visualized by the distinguishability matrices of Fig.~\ref{fig:glass_compression_schemes} in the main text), we are aware of no method other than the distributed IB that can do so while modeling nonlinear relationships with the full expressivity of deep learning.
Finally, we note that the per-particle measurement basis (Fig.~\ref{fig:settransformer} of the main text) similarly has no apparent analogue in terms of Shapley values.
With the distributed IB we are able to inspect the compression channel and study the information allocated to hypothetical particles, even without a well-defined set of features as would be required for analysis by methods that assess feature importance.

\section*{Mutual information bounds}
Bounding mutual information given high-dimensional data is notoriously difficult~\cite{mcallester2020infolimitations,saxe2019}.
Fortunately, there are factors in our favor to facilitate optimization with machine learning and, during evaluation, the recovery of tight bounds on the information transmitted by the compression channels $U_i$.

During training, to optimize the distributed information bottleneck objective requires a lower bound on $I(\boldsymbol{U};Y)$ and an upper bound on $I(U_i;X_i)$.
For $I(\boldsymbol{U};Y)=H(Y)-H(Y|\boldsymbol{U})$, we use the cross entropy loss of the predictions as a lower bound on $H(Y|\boldsymbol{U})$ and ignore $H(Y)$ because it is constant~\cite{alemiVIB2016}.
Regarding $I(U_i;X_i)$, the (distributed) variational information bottleneck objective \cite{alemiVIB2016,aguerriDVIB2021} upper bounds $I(U_i;X_i)$ with the expectation of the Kullback-Leibler (KL) divergence between the encoded distributions $p(u_i|x_i)$ and an arbitrary prior distribution $r(u_i)$ in latent space,
\begin{equation}
    I(U_i;X_i) \le \mathbb{E}_{x_i \sim p(x_i)} [D_\textnormal{KL}(p(u_i|x_i)||r(u_i))].
\end{equation}
Normal distributions are used for both the encoded distribution, $p(u_i|x_i) = \mathcal{N}(\boldsymbol{\mu}=f_\mu(x_i), \boldsymbol{\sigma}=f_\sigma(x_i))$, and the prior, $r(u_i)=\mathcal{N}(\boldsymbol{0}, \boldsymbol{1})$ so that the KL divergence has a simple analytic form.

For evaluation over the course of a training run, the KL divergence is computed for each channel $U_i$ in the process of computing the loss, and could be used for a qualitative sense of information allocation to features~\citep{dib_ml}.
However, the KL divergence is a rather poor estimate of the mutual information, and we seek to know the amount of transmitted information as precisely as possible.
Because the encoded distributions $p(u_i|x_i)$ have a known form, we can use the noise contrastive estimation (InfoNCE) lower bound and ``leave one out'' upper bound from Ref.~\cite{poole2019variational} with a large number of samples to obtain tight bounds on the amount of mutual information in the learned compression schemes (for evaluation).

The lower and upper bounds on $I(U_i;X_i)$ are based on likelihood ratios at points sampled from the dataset $x_i \sim p(x_i)$ and from the corresponding conditional distributions, $u_i \sim p(u_i|x_i)$.
To be specific, the mutual information for each channel $U=f(X)$ (dropping channel indices for simplicity) is lower bounded by 
\begin{equation}\label{eqn:MI_lower}
    I(U;X) \ge \mathbb{E}\left [\frac{1}{K} \sum_i^K  \log \frac{p(u_i|x_i)}{\frac{1}{K} \sum_j^K p(u_i|x_j)} \right ]
\end{equation}
and upper bounded by
\begin{equation}\label{eqn:MI_upper}
    I(U;X) \le \mathbb{E} \left [ \frac{1}{K} \sum_i^K \log \frac{p(u_i|x_i)}{\frac{1}{K-1} \sum_{j\ne i}^K p(u_i|x_j)} \right ].
\end{equation}
The expectation values in both equations are taken over samples $\{u_i,x_i\}_{i=1}^K$ of size $K$ extracted repeatedly from the joint distribution $p(u,x)=p(x)p(u|x)$.
We estimated with as large an evaluation batch size $K$ as feasible given memory and time considerations, and then averaged over multiple batches to reduce the variance of the bound.
Evaluation with a batch size of 1024, averaged over 8 draws, yielded bounds on the mutual information that was on the order of 0.01 bits for the Boolean circuit and glass data.
The size of the validation dataset for the glass and the size of the truth table of the Boolean circuit were both on the order of one thousand points. 
Hence, the benefit of averaging comes from repeated sampling of the latent representations.

\begin{figure*}
    \centering
    \includegraphics[width=\textwidth]{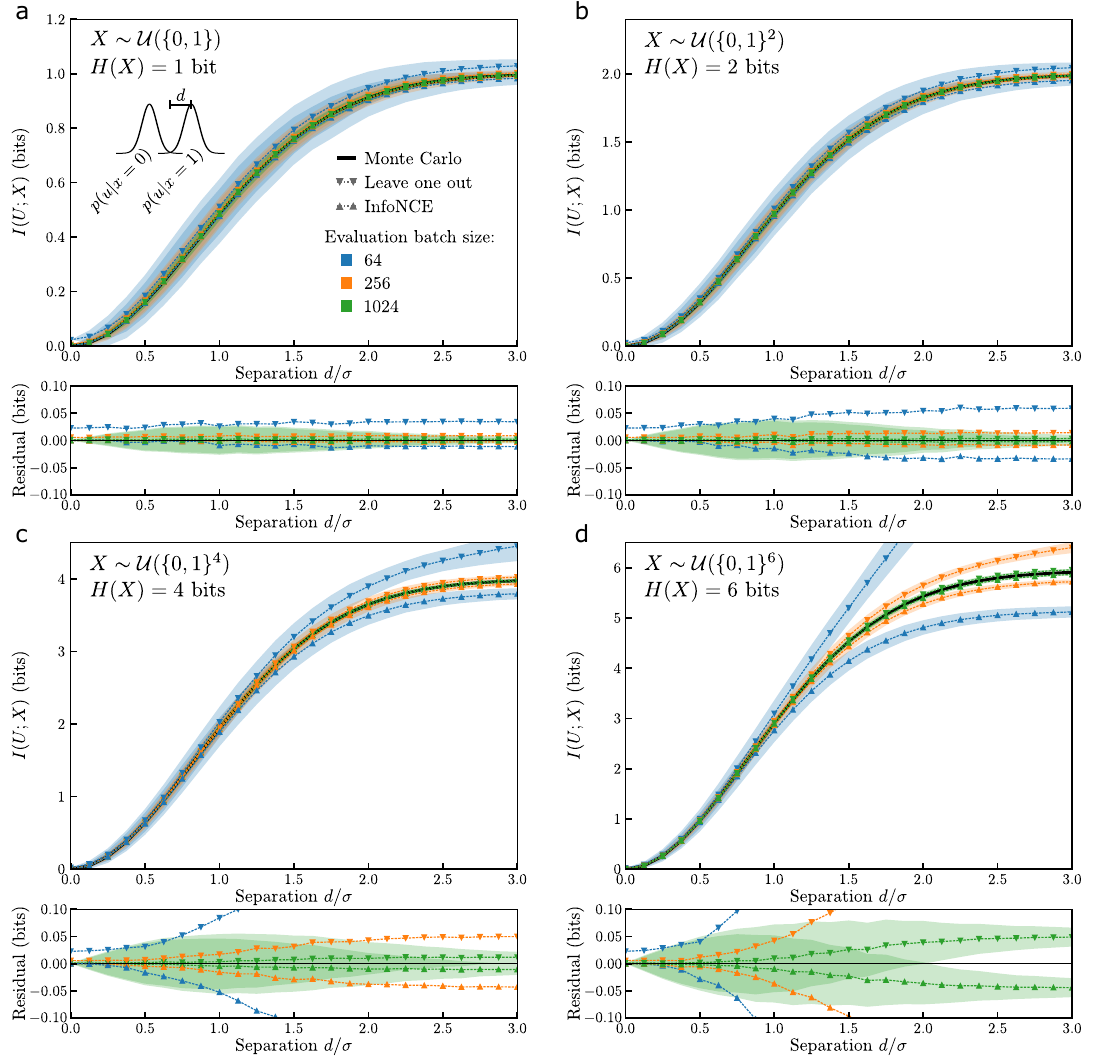}
    \caption{\textbf{Performance of mutual information bounds.}
    A parameterized compression scheme encodes a discrete $X$ to a separate normal distribution per outcome $x$, each with unit variance in $\mathbb{R}^{32}$ and spaced from the origin by a separation distance $d$.
    The expectation and standard deviation of the ``leave one out'' upper and InfoNCE lower bounds are the dashed line and shaded regions, respectively, evaluated over 4096 batches.
    In the residual plots, which show the difference between the bounds and the Monte Carlo estimate, only the standard deviations for the evaluation batch size of 1024 are displayed.
    Variables $X$ with \textbf{(a)} one, \textbf{(b)} two, \textbf{(c)} four, and \textbf{(d)} six bits of entropy are used to generate the dataset for evaluation.
    }
    \label{fig:infobounds}
\end{figure*}

We show in Fig.~\ref{fig:infobounds} the performance of the mutual information bounds for compression schemes that encode up to several bits of information.
$X$ is a discrete random variable that is uniformly distributed over its support and has one to six bits of entropy; for each $X$ a fixed dataset of size 1024 was sampled for mutual information estimation according to the following method of compression.
Each outcome $x$ was encoded to a normal distribution with unit variance in 32-dimensional space, $p(u|x)=\mathcal{N}(\boldsymbol{\mu}, \boldsymbol{1})$.
The encoded distributions were placed along orthogonal axes a distance $d$ from the origin; in the limits of $d=0$ and $d\gg 1$ the information transmitted by the compression scheme is 0 and $H(X)$, respectively.

A Monte Carlo estimate of the mutual information sampled 2$\times$10$^5$ points from $p(u,x)$ to compute $\mathbb{E}_{p(u,x)}[\log p(u|x)/p(u)]$.
The ``leave one out'' upper and InfoNCE lower bounds were computed with different evaluation batch sizes $K$, and averaged over 4096 sampled batches.
The standard deviation of the bounds is displayed as the shaded region around each trace, and is left out of the plots for the residual (the difference between the bound and the Monte Carlo estimate) for all but the evaluation batch size of 1024.

When the information contained in the compression is less than about two bits---as was the case for the majority of the experiments of the main text---the bounds are tight in expectation for even the smallest evaluation batch size.
The variance is reducible by averaging over multiple batches.
As the transmitted information grows, the benefit of increasing the evaluation batch size grows more pronounced, though bounds with a range of less than 0.1 bits can still be achieved for up to six bits of transmitted information. 

\subsection*{Information transmitted per particle}
For the per-particle measurement scheme on the glass data, a single compression channel $U$ was used for all particles.
The information conveyed by the channel $I(U;X)$ may be estimated as above, with $X$ being the particle position and type. Note that we are particularly interested in the information cost for specific particle positions and for each particle type.
The outer summation of the bounds (Eqn.~\ref{eqn:MI_lower} and \ref{eqn:MI_upper}) serves to average over the measurement outcomes $x_i$ in a random sample; we use the summand corresponding to $\{x_i, u_i\}$ as the information contribution for the specific outcome $x_i$.
To generate the information heatmaps of Fig.~\ref{fig:settransformer}b in the main text, we randomly sampled 512 neighborhoods from the dataset, corresponding to an evaluation batch size $K=512 \ \textnormal{neighborhoods} \times50  \ \textnormal{particles / neighborhood} =25,600$ particles (data points), and averaged over 100 such batches. 
A probe particle with specified particle type and position (one for each point in the grid) was inserted into the batch, and then the corresponding summand for the lower and upper information bounds served to quantify the information transmitted per particle.
To be specific, 
\begin{equation}
    I(X=x;U) \ge \mathbb{E} \left [ \log \frac{p(u|x)} {\frac{1}{K} \sum_j^K p(u|x_j)}\right ],
\end{equation}
with the expectation taken over $u \sim p(u|x)$ and samples $\{x_i\}_{i=1}^K\sim \prod_i^K p(x)$.
The upper bound differed only by inclusion of the distribution $p(u|x)$ corresponding to the probe point in the denominator's sum.

\section*{Implementation specifics}
All experiments were implemented in TensorFlow and run on a single computer with a 12 GB GeForce RTX 3060 GPU. Computing mutual information bounds repeatedly throughout an optimization run contributed the most to running time.
Including the information estimation, the Boolean circuit optimization took about half an hour, and the glass experiments took several hours.

\subsection*{Boolean circuit}

Each input may take only one of two values (0 or 1), allowing the encoders to be extremely simple.
Trainable scalars $(\vec{\mu}_i,\textnormal{log}\ \vec{\sigma}_i^2)$ were used to encode $p(u_i|x_i)= \mathcal{N} ((2x_i - 1)\times\vec{\mu}_i, \vec{\sigma}_i^2)$.
The decoder was a multilayer perceptron (MLP) consisting of three fully connected layers with 256 \texttt{Leaky ReLU} units ($\alpha=0.3$) each.
We increased the value of $\beta$ logarithmically from $5\times10^{-4}$ to $5$ in $5\times10^4$ steps, with a batch size of 512 input-output pairs sampled randomly from the entire 1024-element truth table.
We found the information decomposition to be insensitive to annealing duration as long as it was around $5\times10^4$ steps or longer.
The Adam optimizer was used with a learning rate of $10^{-3}$.

\subsection*{Amorphous plasticity}

The simulated glass data comes from Ref.~\cite{richard2020indicators}: 10,000 particles in a two-dimensional cell with Lees-Edwards boundary conditions interact via a Lennard-Jones potential, slightly modified to be twice differentiable~\cite{barbot2018simulations}.
Simple shear was applied with energy minimization after each step of applied strain.
The critical mode was identified as the eigenvector---existing in the $2N$-dimensional configuration space of all the particles' positions---of the Hessian whose eigenvalue crossed zero at the onset of global shear stress decrease.
The particle that was identified as the locus of the rearrangement event had the largest contribution to the critical mode~\cite{richard2020indicators}.

We used data from the gradual quench (``GQ'') and rapid quench (high temperature liquid, ``HTL'') protocols. 
Following Ref.~\cite{schoenholz2016natphys}, we considered only neighborhoods with type A particles (the smaller particles) at the center.
We used all of the events in the dataset: 7,255 for the gradually quenched and 10,178 for the rapidly quenched glasses. 
For each rearrangement event with a type A particle as the locus, we selected at random another region from the same system state with a type A particle at the center to serve as a negative example.
90\% of all rearrangement events with type A particles as the locus were used for the training set and the remaining 10\% were used as the validation set; the regions and specific training and validation splits used in this work can be found on the project webpage.

\subsubsection*{Radial density measurement scheme}
For the radial density measurements (Figs.~\ref{fig:glass},\ref{fig:glass_compression_schemes} of the main text), the local neighborhood of each sample was processed using 50 radial density structure functions for each particle type, evenly spaced over the interval $r=[0.5, 4]$.
Specifically, for particle $i$ at the center and the set of neighboring particles $\mathcal{S}_A$ of type A,
\begin{equation}
    G_A(i;r,\delta)=\sum_{j\in \mathcal{S}_A}\textnormal{exp}(-\frac{(R_{ij}-r)^2}{2\delta^2}),
\end{equation}
\noindent where $R_{ij}$ is the distance between particles $i$ and $j$.
The same expression was used to compute $G_B$, the structure functions for the type B particles in the local neighborhood.
The width parameter $\delta$ was equal to 50\% of each radius interval.

After computing the 100 values summarizing each local neighborhood, the training and validation sets were normalized with the mean and standard deviation of each structure function across the training set.
The best validation results from a logarithmic scan over twenty values from $10^{-3}$ to $10^1$ for the $C$ parameter were used for the value of the SVM accuracy in Fig.~\ref{fig:glass} of the main text.

For the distributed IB, each of the 100 scalar values for the structure functions were input to their own MLP consisting of 2 layers of 128 units with \texttt{tanh} activation.
The embedding dimension of each $U_i$ was 32.
Then the 100 embeddings were concatenated for input to the predictive model, which was an MLP consisting of 3 layers of 256 units with \texttt{tanh} activation.
The output was a single logit to classify whether the particle at the center is the locus of imminent rearrangement.
We increased $\beta$ in equally spaced logarithmic steps from $10^{-6}$ to 1 over 250 epochs (an epoch is one pass through the training data); these hyperparameters were selected based on peak validation performance.
The batch size was 256.
The Adam optimizer was used with a learning rate of $10^{-4}$.

\subsubsection*{Per-particle measurement scheme}
For the per-particle measurements, the nearest 50 particles to the center of each region were compressed by the same encoder, an MLP with two layers of 128 \texttt{Leaky ReLU} activation ($\alpha$=0.1), to a 32-dimensional latent space.
The only information available to the encoder was the particle's position and type, though the values were preprocessed before input to the encoder to help with optimization: for each particle position $\vec{r}=(x,y)$, we concatenated $x^2$, $y^2$, $r=|\vec{r}|$, $\log r$, $\log x^2$, $\log y^2$, and $\vec{r}/r$.
All were positionally encoded (i.e., before being passed to the MLP, inputs were mapped to $x \leftarrow (x, \sin \omega_1 x, \sin \omega_2 x, ... )$) with frequencies $\omega_k = 2^k$, where $k \in {\{1, 2, 3, 4, 5\}}$~\cite{tancik2020fourier,dib_ml}.

After compression, the 50 representations (one for each particle) were input to a set transformer~\cite{lee2019set}, a permutation-invariant architecture that is free to learn how to relate different particles via self-attention.
We used 6 multi-head attention (MHA) blocks with 12 heads each, and a key dimension of 128.
Following Ref.~\cite{lee2019set}, each MHA block adds the output of multi-head attention to a skip connection of the block's input, and applies layer normalization to the sum.
This intermediate output is passed through an MLP (a single layer with 128 \texttt{ReLU} units, in our case) and added to itself (another skip connection) before a second round of layer normalization.
After the MHA blocks, the 50 particle representations were mean-pooled and passed through a final fully connected layer of 256 units with \texttt{Leaky ReLU} activation ($\alpha$=0.1) before outputting a logit for prediction.

Training proceeded for 25,000 training steps, and the learning rate was ramped linearly from zero to $10^{-4}$ over the first 10\% of training.
Over the duration of training, $\beta$ increased logarithmically from $3\times 10^{-8}$ to $3 \times 10^{-3}$.  
The batch size was 64.

\section*{Citation Diversity Statement}
Science is a human endeavour and consequently vulnerable to many forms of bias; the responsible scientist identifies and mitigates such bias wherever possible.
Meta-analyses of research in multiple fields have measured significant bias in how research works are cited, to the detriment of scholars in minority groups~\cite{maliniak2013gender,caplar2017quantitative,chakravartty2018communicationsowhite,dion2018gendered,dworkin2020extent}.
We use this space to amplify studies, perspectives, and tools that we found influential during the execution of this research~\cite{zurn2020citation,dworkin2020citing,zhou2020gender,budrikis2020growing}.

\section*{Glass dataset}
\textbf{glass\_data.tar.gz}
The train and validation splits of the glass data, consisting of local neighborhoods immediately preceding rearrangement events, derived from Ref.~\cite{richard2020indicators}, have been deposited on Figshare as a directory of \texttt{numpy} arrays (\href{https://doi.org/10.6084/m9.figshare.24585150.v1}{https://doi.org/10.6084/m9.figshare.24585150.v1})\cite{glass_dataset}.
The neighborhoods were subsequently ``measured'' as radial densities (Figs.~\ref{fig:glass} and \ref{fig:glass_compression_schemes}) or as per-particle descriptors (Fig.~\ref{fig:settransformer}); the code to process the neighborhoods is in
\href{https://github.com/distributed-information-bottleneck/distributed-information-bottleneck.github.io/tree/main/complex_systems}{the project Github repository}.
Also included in the dataset are the computed radial distribution functions, $g_\text{AA}(r)$ and $g_\text{AB}(r)$, for each quench.


\begin{thebibliography}{10}

\bibitem{nicolis2012foundations}
G Nicolis, C Nicolis, {\em Foundations of complex systems: emergence,
  information and predicition}.
\newblock (World Scientific), (2012).

\bibitem{kwapien2012complexity}
J Kwapień, S Drożdż, Physical approach to complex systems.
\newblock {\em\protect\JournalTitle{Physics Reports}} \textbf{515}, 115--226
  (2012).

\bibitem{mitchell2009complexitybook}
M Mitchell, {\em Complexity: A guided tour}.
\newblock (Oxford University Press), (2009).

\bibitem{newman2011complex}
ME Newman, Complex systems: A survey.
\newblock {\em\protect\JournalTitle{arXiv preprint arXiv:1112.1440}} (2011).

\bibitem{matsuda2000physical}
H Matsuda, Physical nature of higher-order mutual information: Intrinsic
  correlations and frustration.
\newblock {\em\protect\JournalTitle{Physical review E}} \textbf{62}, 3096
  (2000).

\bibitem{koch2018natphys}
M Koch-Janusz, Z Ringel, Mutual information, neural networks and the
  renormalization group.
\newblock {\em\protect\JournalTitle{Nature Physics}} \textbf{14}, 578--582
  (2018).

\bibitem{grassberger1986toward}
P Grassberger, Toward a quantitative theory of self-generated complexity.
\newblock {\em\protect\JournalTitle{International Journal of Theoretical
  Physics}} \textbf{25}, 907--938 (1986).

\bibitem{tononi1994measure}
G Tononi, O Sporns, GM Edelman, A measure for brain complexity: relating
  functional segregation and integration in the nervous system.
\newblock {\em\protect\JournalTitle{Proceedings of the National Academy of
  Sciences}} \textbf{91}, 5033--5037 (1994).

\bibitem{gellmann1996effective}
M Gell-Mann, S Lloyd, Information measures, effective complexity, and total
  information.
\newblock {\em\protect\JournalTitle{Complexity}} \textbf{2}, 44--52 (1996).

\bibitem{rosas2019Oinfo}
FE Rosas, PA Mediano, M Gastpar, HJ Jensen, Quantifying high-order
  interdependencies via multivariate extensions of the mutual information.
\newblock {\em\protect\JournalTitle{Physical Review E}} \textbf{100}, 032305
  (2019).

\bibitem{golan2022pnas}
A Golan, J Harte, Information theory: A foundation for complexity science.
\newblock {\em\protect\JournalTitle{Proceedings of the National Academy of
  Sciences}} \textbf{119}, e2119089119 (2022).

\bibitem{lecun2015deep}
Y LeCun, Y Bengio, G Hinton, Deep learning.
\newblock {\em\protect\JournalTitle{nature}} \textbf{521}, 436--444 (2015).

\bibitem{rudin2019stop}
C Rudin, Stop explaining black box machine learning models for high stakes
  decisions and use interpretable models instead.
\newblock {\em\protect\JournalTitle{Nature Machine Intelligence}} \textbf{1},
  206--215 (2019).

\bibitem{rudin2022interpretable}
C Rudin, et~al., Interpretable machine learning: Fundamental principles and 10
  grand challenges.
\newblock {\em\protect\JournalTitle{Statistics Surveys}} \textbf{16}, 1--85
  (2022).

\bibitem{molnar2022interpretableML}
C Molnar, {\em Interpretable Machine Learning: A Guide for Making Black Box
  Models Explainable}.
\newblock (2022).

\bibitem{dib_ml}
KA Murphy, DS Bassett, Interpretability with full complexity by constraining
  feature information in {\em International Conference on Learning
  Representations ({ICLR})}.
\newblock (2023).

\bibitem{aguerri2018DIB}
I Estella~Aguerri, A Zaidi, Distributed information bottleneck method for
  discrete and gaussian sources in {\em International Zurich Seminar on
  Information and Communication (IZS 2018) Proceedings}.
\newblock (ETH Zurich), pp. 35--39 (2018).

\bibitem{aguerriDVIB2021}
IE Aguerri, A Zaidi, Distributed variational representation learning.
\newblock {\em\protect\JournalTitle{IEEE Transactions on Pattern Analysis and
  Machine Intelligence}} \textbf{43}, 120--138 (2021).

\bibitem{tishbyIB2000}
N Tishby, FC Pereira, W Bialek, The information bottleneck method.
\newblock {\em\protect\JournalTitle{arXiv preprint physics/0004057}} (2000).

\bibitem{savage1998models}
JE Savage, {\em Models of computation}.
\newblock (Addison-Wesley Reading, MA) Vol.{} 136, (1998).

\bibitem{chaves2006methods}
M Chaves, ED Sontag, R Albert, Methods of robustness analysis for {B}oolean
  models of gene control networks.
\newblock {\em\protect\JournalTitle{IEE Proceedings-Systems Biology}}
  \textbf{153}, 154--167 (2006).

\bibitem{huynh2019gene}
VA Huynh-Thu, G Sanguinetti, Gene regulatory network inference: an introductory
  survey in {\em Gene Regulatory Networks}.
\newblock (Springer), pp. 1--23 (2019).

\bibitem{cubuk2017science}
ED Cubuk, et~al., Structure-property relationships from universal signatures of
  plasticity in disordered solids.
\newblock {\em\protect\JournalTitle{Science}} \textbf{358}, 1033--1037 (2017).

\bibitem{murphy2019transforming}
KA Murphy, KA Dahmen, HM Jaeger, Transforming mesoscale granular plasticity
  through particle shape.
\newblock {\em\protect\JournalTitle{Physical Review X}} \textbf{9}, 011014
  (2019).

\bibitem{jaeger1992}
HM Jaeger, SR Nagel, Physics of the granular state.
\newblock {\em\protect\JournalTitle{Science}} \textbf{255}, 1523--1531 (1992).

\bibitem{liu1998jammingcool}
AJ Liu, SR Nagel, Jamming is not just cool any more.
\newblock {\em\protect\JournalTitle{Nature}} \textbf{396}, 21--22 (1998).

\bibitem{bi2011jammingshear}
D Bi, J Zhang, B Chakraborty, RP Behringer, Jamming by shear.
\newblock {\em\protect\JournalTitle{Nature}} \textbf{480}, 355--358 (2011).

\bibitem{berthier2019gardner}
L Berthier, et~al., Gardner physics in amorphous solids and beyond.
\newblock {\em\protect\JournalTitle{The Journal of chemical physics}}
  \textbf{151}, 010901 (2019).

\bibitem{cover1999elements}
TM Cover, JA Thomas, {\em Elements of information theory}.
\newblock (John Wiley \& Sons), (1999).

\bibitem{shannon1948mathematical}
CE Shannon, A mathematical theory of communication.
\newblock {\em\protect\JournalTitle{The Bell System Technical Journal}}
  \textbf{27}, 379--423 (1948).

\bibitem{steiner2021distributedcompression}
S Steiner, V Kuehn, Distributed compression using the information bottleneck
  principle in {\em ICC 2021 - IEEE International Conference on
  Communications}.
\newblock pp. 1--6 (2021).

\bibitem{alemiVIB2016}
AA Alemi, I Fischer, JV Dillon, K Murphy, Deep variational information
  bottleneck.
\newblock {\em\protect\JournalTitle{International Conference on Learning
  Representations ({ICLR})}} (2017).

\bibitem{betavae}
I Higgins, et~al., $\beta$-vae: Learning basic visual concepts with a
  constrained variational framework.
\newblock {\em\protect\JournalTitle{International Conference on Learning
  Representations (ICLR)}} (2017).

\bibitem{poole2019variational}
B Poole, S Ozair, A Van Den~Oord, A Alemi, G Tucker, On variational bounds of
  mutual information in {\em International Conference on Machine Learning}.
\newblock (PMLR), pp. 5171--5180 (2019).

\bibitem{mcallester2020infolimitations}
D McAllester, K Stratos, Formal limitations on the measurement of mutual
  information in {\em International Conference on Artificial Intelligence and
  Statistics}.
\newblock (PMLR), pp. 875--884 (2020).

\bibitem{shap}
SM Lundberg, SI Lee, A unified approach to interpreting model predictions in
  {\em Advances in Neural Information Processing Systems 30}, eds.{} I Guyon,
  et~al.
\newblock (Curran Associates, Inc.), pp. 4765--4774 (2017).

\bibitem{sage}
I Covert, SM Lundberg, SI Lee, Understanding global feature contributions with
  additive importance measures.
\newblock {\em\protect\JournalTitle{Advances in Neural Information Processing
  Systems}} \textbf{33}, 17212--17223 (2020).

\bibitem{argon1979bubbles}
AS Argon, HY Kuo, Plastic flow in a disordered bubble raft (an analog of a
  metallic glass).
\newblock {\em\protect\JournalTitle{Materials Science and Engineering}}
  \textbf{39}, 101--109 (1979).

\bibitem{falk1998dynamics}
ML Falk, JS Langer, Dynamics of viscoplastic deformation in amorphous solids.
\newblock {\em\protect\JournalTitle{Physical Review E}} \textbf{57}, 7192
  (1998).

\bibitem{manning2011vibrational}
ML Manning, AJ Liu, Vibrational modes identify soft spots in a sheared
  disordered packing.
\newblock {\em\protect\JournalTitle{Physical Review Letters}} \textbf{107},
  108302 (2011).

\bibitem{dahmen2011simple}
KA Dahmen, Y Ben-Zion, JT Uhl, A simple analytic theory for the statistics of
  avalanches in sheared granular materials.
\newblock {\em\protect\JournalTitle{Nature Physics}} \textbf{7}, 554--557
  (2011).

\bibitem{richard2020indicators}
D Richard, et~al., Predicting plasticity in disordered solids from structural
  indicators.
\newblock {\em\protect\JournalTitle{Physical Review Materials}} \textbf{4},
  113609 (2020).

\bibitem{dunleavy2012information}
AJ Dunleavy, K Wiesner, CP Royall, Using mutual information to measure order in
  model glass formers.
\newblock {\em\protect\JournalTitle{Physical Review E}} \textbf{86}, 041505
  (2012).

\bibitem{jack2014information}
RL Jack, AJ Dunleavy, CP Royall, Information-theoretic measurements of coupling
  between structure and dynamics in glass formers.
\newblock {\em\protect\JournalTitle{Physical review letters}} \textbf{113},
  095703 (2014).

\bibitem{dunleavy2015mutual}
AJ Dunleavy, K Wiesner, R Yamamoto, CP Royall, Mutual information reveals
  multiple structural relaxation mechanisms in a model glass former.
\newblock {\em\protect\JournalTitle{Nature communications}} \textbf{6}, 6089
  (2015).

\bibitem{behler2007structurefns}
J Behler, M Parrinello, Generalized neural-network representation of
  high-dimensional potential-energy surfaces.
\newblock {\em\protect\JournalTitle{Physical Review Letters}} \textbf{98},
  146401 (2007).

\bibitem{cubuk2015PRL}
ED Cubuk, et~al., Identifying structural flow defects in disordered solids
  using machine-learning methods.
\newblock {\em\protect\JournalTitle{Physical Review Letters}} \textbf{114},
  108001 (2015).

\bibitem{schoenholz2016natphys}
SS Schoenholz, ED Cubuk, DM Sussman, E Kaxiras, AJ Liu, A structural approach
  to relaxation in glassy liquids.
\newblock {\em\protect\JournalTitle{Nature Physics}} \textbf{12}, 469--471
  (2016).

\bibitem{softnessGrainBoundaries}
TA Sharp, et~al., Machine learning determination of atomic dynamics at grain
  boundaries.
\newblock {\em\protect\JournalTitle{Proceedings of the National Academy of
  Sciences}} \textbf{115}, 10943--10947 (2018).

\bibitem{softnessFilms}
DM Sussman, SS Schoenholz, ED Cubuk, AJ Liu, Disconnecting structure and
  dynamics in glassy thin films.
\newblock {\em\protect\JournalTitle{Proceedings of the National Academy of
  Sciences}} \textbf{114}, 10601--10605 (2017).

\bibitem{ridout2021avalanche}
G Zhang, SA Ridout, AJ Liu, Interplay of rearrangements, strain, and local
  structure during avalanche propagation.
\newblock {\em\protect\JournalTitle{Physical Review X}} \textbf{11}, 041019
  (2021).

\bibitem{dib_orig}
KA Murphy, DS Bassett, The distributed information bottleneck reveals the
  explanatory structure of complex systems.
\newblock {\em\protect\JournalTitle{arXiv preprint arXiv:2204.07576}} (2022).

\bibitem{lee2019set}
J Lee, et~al., Set transformer: A framework for attention-based
  permutation-invariant neural networks in {\em International conference on
  machine learning}.
\newblock (PMLR), pp. 3744--3753 (2019).

\bibitem{saxe2019}
AM Saxe, et~al., On the information bottleneck theory of deep learning.
\newblock {\em\protect\JournalTitle{Journal of Statistical Mechanics: Theory
  and Experiment}} \textbf{2019}, 124020 (2019).

\bibitem{williams2010PID}
PL Williams, RD Beer, Nonnegative decomposition of multivariate information.
\newblock {\em\protect\JournalTitle{arXiv preprint arXiv:1004.2515}} (2010).

\bibitem{zaidi2020IBreview}
A Zaidi, I Estella-Aguerri, S Shamai~(Shitz), On the information bottleneck
  problems: Models, connections, applications and information theoretic views.
\newblock {\em\protect\JournalTitle{Entropy}} \textbf{22} (2020).

\bibitem{goldfeld2020information}
Z Goldfeld, Y Polyanskiy, The information bottleneck problem and its
  applications in machine learning.
\newblock {\em\protect\JournalTitle{IEEE Journal on Selected Areas in
  Information Theory}} \textbf{1}, 19--38 (2020).

\bibitem{gordonrelevance2021}
A Gordon, A Banerjee, M Koch-Janusz, Z Ringel, Relevance in the renormalization
  group and in information theory.
\newblock {\em\protect\JournalTitle{Physical Review Letters}} \textbf{126},
  240601 (2021).

\bibitem{kline2021RGIB}
AG Kline, S Palmer, Gaussian information bottleneck and the non-perturbative
  renormalization group.
\newblock {\em\protect\JournalTitle{New Journal of Physics}} (2021).

\bibitem{wang2019PIB}
Y Wang, JML Ribeiro, P Tiwary, Past--future information bottleneck for sampling
  molecular reaction coordinate simultaneously with thermodynamics and
  kinetics.
\newblock {\em\protect\JournalTitle{Nature Communications}} \textbf{10}, 1--8
  (2019).

\bibitem{kolchinsky2018caveats}
A Kolchinsky, BD Tracey, S Van~Kuyk, Caveats for information bottleneck in
  deterministic scenarios.
\newblock {\em\protect\JournalTitle{International Conference on Learning
  Representations (ICLR)}} (2019).

\bibitem{kolchinskyNonlinearIB2019}
A Kolchinsky, BD Tracey, DH Wolpert, Nonlinear information bottleneck.
\newblock {\em\protect\JournalTitle{Entropy}} \textbf{21} (2019).

\bibitem{gutknecht2021bits}
AJ Gutknecht, M Wibral, A Makkeh, Bits and pieces: Understanding information
  decomposition from part-whole relationships and formal logic.
\newblock {\em\protect\JournalTitle{Proceedings of the Royal Society A}}
  \textbf{477}, 20210110 (2021).

\bibitem{kolchinsky2022PID}
A Kolchinsky, A novel approach to the partial information decomposition.
\newblock {\em\protect\JournalTitle{Entropy}} \textbf{24}, 403 (2022).

\bibitem{varley2022emergence}
TF Varley, E Hoel, Emergence as the conversion of information: A unifying
  theory.
\newblock {\em\protect\JournalTitle{Philosophical Transactions of the Royal
  Society A}} \textbf{380}, 20210150 (2022).

\bibitem{ehrlich2022partial}
DA Ehrlich, AC Schneider, M Wibral, V Priesemann, A Makkeh, Partial information
  decomposition reveals the structure of neural representations.
\newblock {\em\protect\JournalTitle{arXiv preprint arXiv:2209.10438}} (2022).

\bibitem{timme2014SynRedReview}
N Timme, W Alford, B Flecker, JM Beggs, Synergy, redundancy, and multivariate
  information measures: an experimentalist’s perspective.
\newblock {\em\protect\JournalTitle{Journal of computational neuroscience}}
  \textbf{36}, 119--140 (2014).

\bibitem{glass_dataset}
KA Murphy, DS Bassett, Regions of simulated glasses for binary classification
  about imminent rearrangement https://doi.org/10.6084/m9.figshare.24585150.v1
  (2023).

\bibitem{barbot2018simulations}
A Barbot, et~al., Local yield stress statistics in model amorphous solids.
\newblock {\em\protect\JournalTitle{Physical Review E}} \textbf{97}, 033001
  (2018).

\bibitem{tancik2020fourier}
M Tancik, et~al., Fourier features let networks learn high frequency functions
  in low dimensional domains.
\newblock {\em\protect\JournalTitle{Advances in Neural Information Processing
  Systems}} \textbf{33}, 7537--7547 (2020).

\bibitem{maliniak2013gender}
D Maliniak, R Powers, BF Walter, The gender citation gap in international
  relations.
\newblock {\em\protect\JournalTitle{International Organization}} \textbf{67},
  889--922 (2013).

\bibitem{caplar2017quantitative}
N Caplar, S Tacchella, S Birrer, Quantitative evaluation of gender bias in
  astronomical publications from citation counts.
\newblock {\em\protect\JournalTitle{Nature Astronomy}} \textbf{1}, 1--5 (2017).

\bibitem{chakravartty2018communicationsowhite}
P Chakravartty, R Kuo, V Grubbs, C McIlwain, {\#CommunicationSoWhite}.
\newblock {\em\protect\JournalTitle{Journal of Communication}} \textbf{68},
  254--266 (2018).

\bibitem{dion2018gendered}
ML Dion, JL Sumner, SM Mitchell, Gendered citation patterns across political
  science and social science methodology fields.
\newblock {\em\protect\JournalTitle{Political Analysis}} \textbf{26}, 312--327
  (2018).

\bibitem{dworkin2020extent}
JD Dworkin, et~al., The extent and drivers of gender imbalance in neuroscience
  reference lists.
\newblock {\em\protect\JournalTitle{Nature Neuroscience}} \textbf{23}, 918--926
  (2020).

\bibitem{zurn2020citation}
P Zurn, DS Bassett, NC Rust, The citation diversity statement: a practice of
  transparency, a way of life.
\newblock {\em\protect\JournalTitle{Trends in Cognitive Sciences}} \textbf{24},
  669--672 (2020).

\bibitem{dworkin2020citing}
J Dworkin, P Zurn, DS Bassett, {(In)}citing action to realize an equitable
  future.
\newblock {\em\protect\JournalTitle{Neuron}} \textbf{106}, 890--894 (2020).

\bibitem{zhou2020gender}
D Zhou, et~al., Gender diversity statement and code notebook v1. 0.
\newblock {\em\protect\JournalTitle{Zenodo}} (2020).

\bibitem{budrikis2020growing}
Z Budrikis, Growing citation gender gap.
\newblock {\em\protect\JournalTitle{Nature Reviews Physics}} \textbf{2},
  346--346 (2020).

\end{thebibliography}
\end{document}